\documentclass{article}

 \usepackage[preprint]{neurips_2025}
 \usepackage{graphicx}
 \usepackage{subcaption}
 \usepackage{amsmath}
 \usepackage{algorithm}
 \usepackage{algpseudocode}
 \usepackage[export]{adjustbox}
 \usepackage{subcaption}
 \usepackage{hyperref}

\usepackage[utf8]{inputenc} 
\usepackage[T1]{fontenc}    
\usepackage{hyperref}       
\usepackage{url}            
\usepackage{booktabs}       
\usepackage{amsfonts}       
\usepackage{nicefrac}       
\usepackage{microtype}      
\usepackage{xcolor} 
\usepackage{wrapfig}       

\title{Laplacian Score Sharpening for Mitigating Hallucination in Diffusion Models}

\author{%
 Barath Chandran.C$^1$, Srinivas Anumasa$^2$, Dianbo Liu$^2$ \\ \textsuperscript{1}Indian Institute of Technology, Roorkee
   \textsuperscript{2}National University of Singapore  \\
  \texttt{barath\_cc@ece.iitr.ac.in} \\
}

\begin{document}

\maketitle

\begin{abstract}

    Diffusion models, though successful, are known to suffer from hallucinations that create incoherent or unrealistic samples. Recent works have attributed this to the phenomenon of mode interpolation and score smoothening, but they lack a method to prevent their generation during sampling. In this paper, we propose a post-hoc adjustment to the score function during inference that leverages the Laplacian (or sharpness) of the score to reduce mode interpolation hallucination in unconditional diffusion models across 1D, 2D, and high-dimensional image data. We derive an efficient Laplacian approximation for higher dimensions using a finite-difference variant of the Hutchinson trace estimator. We show that this correction significantly reduces the rate of hallucinated samples across toy 1D/2D distributions and a high-dimensional image dataset. Furthermore, our analysis explores the relationship between the Laplacian and uncertainty in the score.
\end{abstract}

\section{Introduction}

\begin{wrapfigure}{r}{0.33\textwidth}
    \vspace{-0.2cm}
    \centering
    \includegraphics[width=0.98\linewidth]{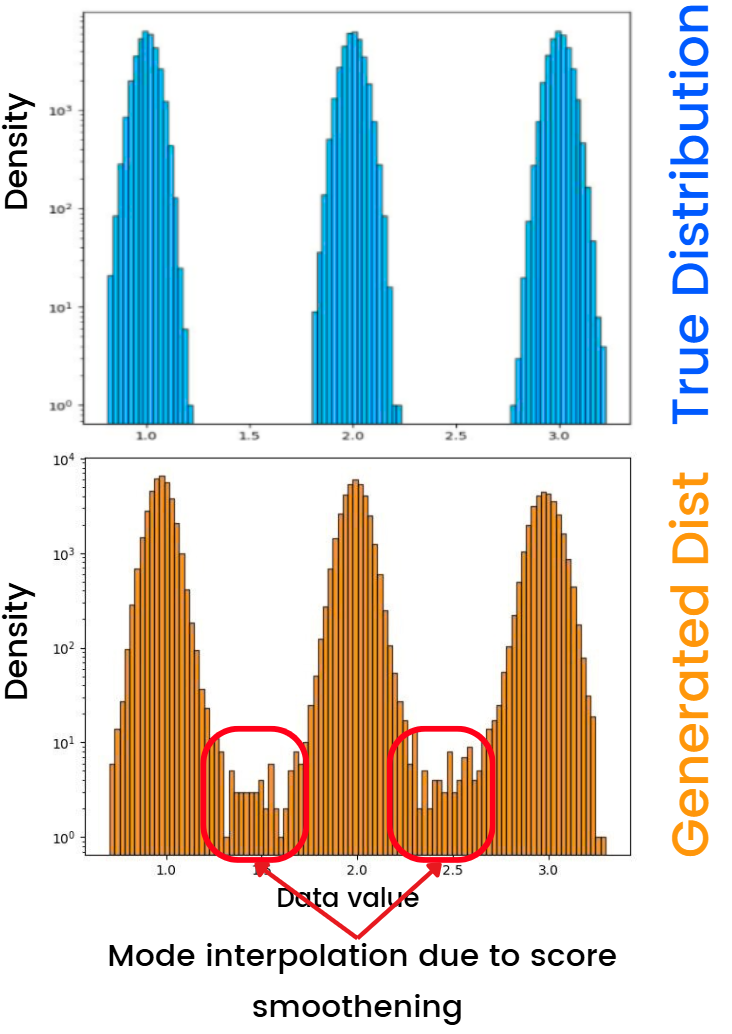}
    \caption{Illustration of the mode interpolation phenomenon in diffusion models, generated samples fall in between the true data modes, resulting in hallucinations}
    \label{fig:illustration}
\end{wrapfigure}

Generative modeling has advanced rapidly, enabling the creation of highly realistic and complex data across diverse fields such as image synthesis, language modeling, and molecular design. Among these approaches, diffusion models have gained prominence by generating samples through the iterative refinement of noisy data \cite{DDPM}. This refinement is guided by the score function—the gradient of the log-probability—which determines the noise to be removed at each timestep and thereby reveals the underlying data distribution \cite{Yangsong}. Despite their success, these models still suffer from artifacts and hallucinations, where generated samples contain unrealistic or incoherent elements not present in the training distribution. A common example is the generation of human hands with extra fingers, illustrating a failure to perfectly capture the true data manifold.

A key study by Aithal et al. \cite{aithal2024understandinghallucinationsdiffusionmodels} identifies mode interpolation as the primary source of such hallucinations in unconditional image generation as shown in Figure ~\ref{fig:illustration}, attributing it to an excessive smoothing of the score function in inter-mode regions. This work directly motivates our goal of targeting these regions of uncertain score estimates. In a complementary work, Jeon et al. \cite{jeon2025understandingmitigatingmemorizationgenerative} analyze the curvature of the log-probability (the trace of the score's first derivative) to quantify overfitting and mitigate memorization of training samples in diffusion models. This concept of leveraging curvature is supported by Lee and Park \cite{lee2023explicitcurvatureregularizationdeep}, who introduced curvature-based regularization using Hutchinson’s trace estimator \cite{hutchinson1990stochastic} for efficient training of Autoencoders. While the memorization study \cite{jeon2025understandingmitigatingmemorizationgenerative} focuses on the score and its Jacobian (first derivative), we hypothesize that the Laplacian (the trace of the Hessian, or second derivative) provides a crucial signal for uncertainty in the score function. This leads us to ask:

\begin{itemize}
    \item Can the Laplacian be leveraged to enhance the score in inter-mode regions, thereby mitigating hallucinations?
    \item Can the Laplacian also serve as an indicator of regions where the model exhibits uncertainty in its estimated score?
\end{itemize}

\section{Background}

The core of the diffusion process is guided by the score function \cite{song2020generativemodelingestimatinggradients}, defined as the gradient of the log-probability density, $\nabla_x \log p(x)$. This vector field points in the direction of steepest ascent in the data probability density. During sampling, this effectively acts as a guide, indicating the path from noisy inputs toward cleaner data. In practice, the model learns to estimate this score through the denoising objective of the diffusion process. A key result \cite{Yangsong} establishes that the noise $\epsilon$ predicted by a DDPM is a scaled estimate of this score for the noisy distribution at timestep $t$: 
\[
\epsilon_\theta(x_t, t) \propto -\,\nabla_{x_t} \log p_t(x_t)
\]
The Jacobian(first derivative) of the score function is equivalent to the Hessian of the log-probability, $\nabla^2 \log p(\mathbf{x})$. The properties of this matrix denote the local curvature of the data distribution. Regions with high curvature (e.g., a large negative trace) indicate high local confidence, while regions with low curvature (trace near zero) indicate low confidence \cite{jeon2025understandingmitigatingmemorizationgenerative}. This is clearly observed when a generated sample $\mathbf{x}$ is near a training sample where the log-probability surface is sharply peaked, resulting in a large negative trace of high magnitude. Consequently, the trace has been proposed as a regularizer to avoid memorization of training samples.

\begin{figure}[h!]
    \centering
    \begin{subfigure}[b]{0.44\linewidth}
        \includegraphics[width=\linewidth]{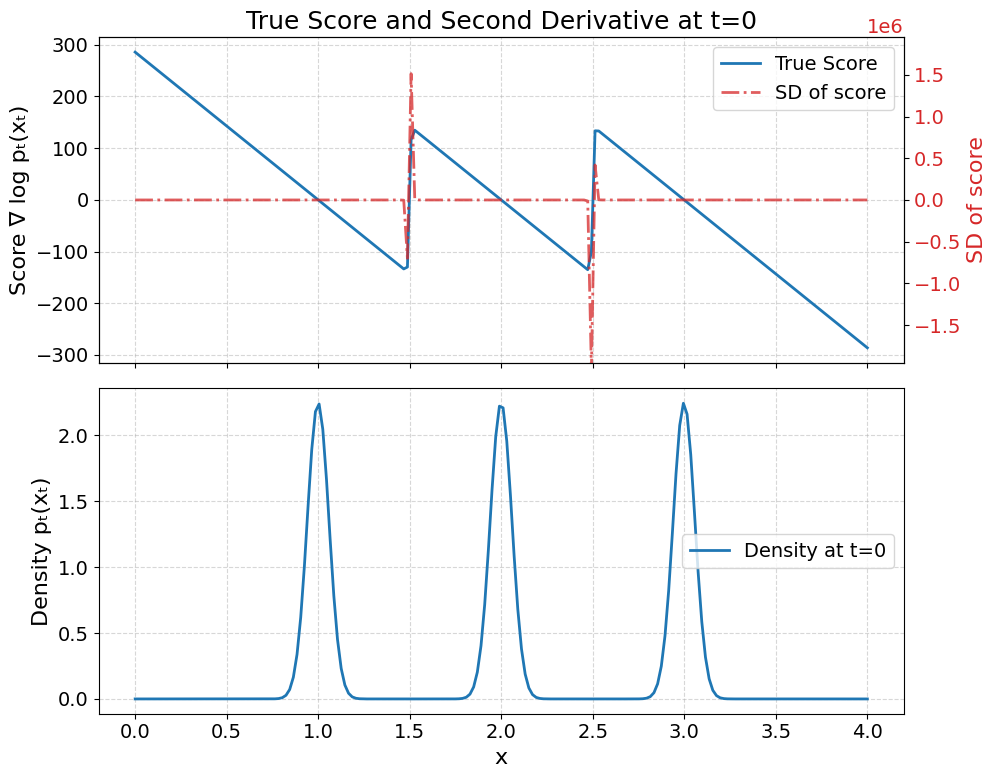}
        \caption{}
        \label{fig:True_score}
    \end{subfigure}
    \begin{subfigure}[b]{0.44\linewidth}
        \includegraphics[width=\linewidth]{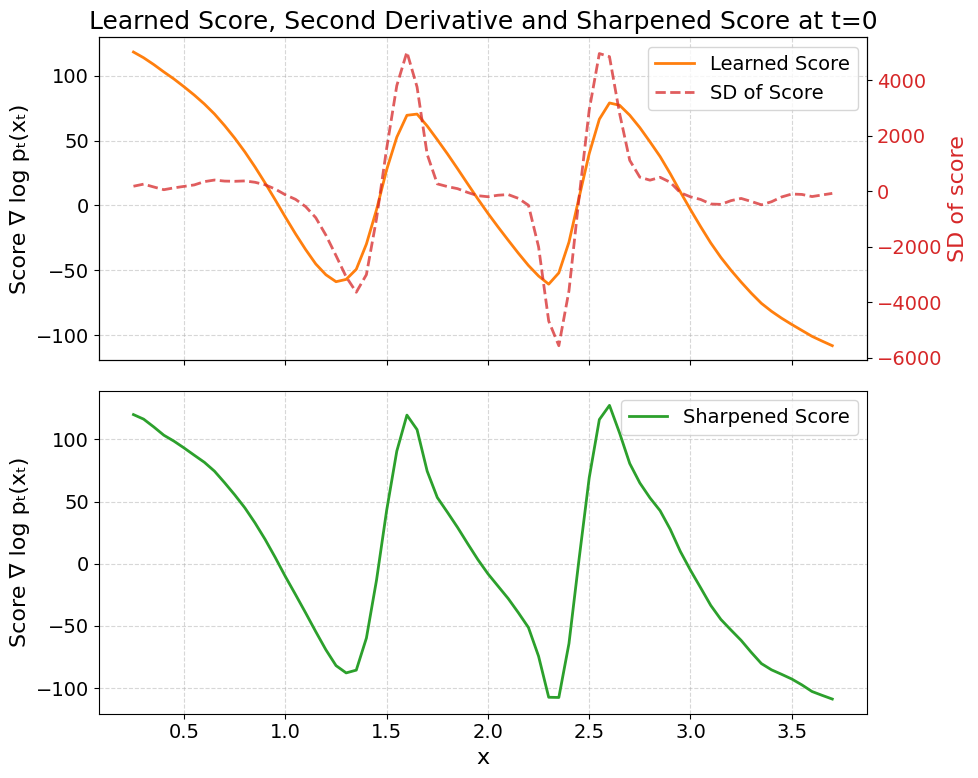}
        \caption{}
        \label{fig:Learned_score}
    \end{subfigure}
    \caption{(a) A toy 1D distribution with modes at $x=\{1,2,3\}$ along with its true score function (blue) and the second derivative of the true score. 
    (b) A visualization of score sharpening in the 1D scenario, from top to bottom: the learned score function (orange) corresponding to the distribution in (a), the second derivative (Laplacian) of the learned score, and the new sharpened score function (green).}
    \label{fig:thriple_combined}
\end{figure}

Building on this, we hypothesize that the second derivative of the score captures abrupt changes in the confidence—highlighting regions where the model is uncertain about the correct trajectory. As shown in Figure \ref{fig:thriple_combined}, both the true and learned scores exhibit non-zero second-order derivatives primarily in inter-mode regions, while remaining near zero at the modes themselves. However, the learned score is overly smoothed due to the model’s inability to represent sharp transitions \cite{aithal2024understandinghallucinationsdiffusionmodels}, resulting in a diminished signal near inter-modes and causing sample trajectories to stagnate. To address this, we aim to sharpen the score, selectively increasing its magnitude in these uncertain regions to guide sample trajectories away from them and mitigate mode-interpolation hallucinations. 

\section{Methodology}

\subsection{Peak Sharpening with Second-Order Derivative}

Building on this observation, peak sharpening directly leverages the second derivative to enhance the peaks of the score function. Specifically, we perform a weighted subtraction of the original function and its second derivative: 
\begin{equation}
R_j = Y_j - \alpha Y_j''
\label{eq:r_j}
\end{equation}
where \( R_j \) is the sharpened (peak-enhanced) score, \( Y_j \) is the original score, and \( Y_j'' \) denotes its second derivative.

 To reduce computational overhead, we approximate the Second order derivative using the finite difference method at various points. This approximation can be seamlessly integrated into the sampling process of a DDPM (or, more broadly, any score-based diffusion model) to improve sample quality and mitigate hallucinations. Figure \ref{fig:Learned_score} illustrates the effect of the sharpening operation on a 1D score function. After peak enhancement, the score function (green) exhibits sharper transitions at inter-modes compared to the learned score (orange), while remaining unchanged elsewhere. Algorithm \ref{alg:score_sharpening_1d} outlines the sampling procedure for the 1D scenario with finite difference approximation for the second derivative. For 1D (and 2D setting) the score sharpening algorithm involves three hyperparameters:

\begin{itemize}
    \item \textbf{Timestep threshold} ($t_{\text{threshold}}$): The forward timestep defining the start of sharpening in the reverse process. A higher $t_{\text{threshold}}$ means sharpening is applied for \textit{more} reverse steps, including \textit{noisier} states. This can improve peak recovery but increases artifact risk. Sharpening is ineffective in very noisy states (high forward $t$), as the score lacks sharp peaks at the inter-modes.(Appendix~\ref{app:time})
    \item \textbf{Perturbation size} ($\delta$):      The finite-difference step size used to estimate the second derivative. Smaller $\delta$ values produce sharper but noisier estimates, while larger $\delta$ values yield smoother but less precise curvature information.
    \item \textbf{Enhancement strength} ($\alpha$):       The scaling factor applied to the negative second derivative during sharpening. Lower $\alpha$ values result in minimal sharpening, whereas higher $\alpha$ values can exaggerate peaks and introduce artifacts.
\end{itemize}

\begin{algorithm}
\caption{Sampling with Score Sharpening in 1D setting}
\label{alg:score_sharpening_1d}
\begin{algorithmic}[1]
\Function{Sharpened\_Denoise}{$x, t, \text{denoise\_fn}, \delta, \alpha, t_{\text{threshold}}$}
    \State $f_x \gets \text{denoise\_fn}(x, t)$ \Comment{$f_x$ estimates noise $\epsilon$ at timestep $t$}
    \If{$t < t_{\text{threshold}}$} \Comment{$t_{\text{threshold}}$: timestep threshold}
        \State $f_x^{+} \gets \text{denoise\_fn}(x + \delta, t)$ \Comment{$\delta$: perturbation size}
        \State $f_x^{-} \gets \text{denoise\_fn}(x - \delta, t)$
        \State $\text{second\_derivative} \gets \frac{ f_x^{+} + f_x^{-} -2 f_x}{\delta^2}$
        \State $f_x \gets f_x - \alpha \cdot \text{second\_derivative}$ \Comment{$\alpha$: sharpening strength}
    \EndIf
    \State \Return $f_x$
\EndFunction
\Statex
\State \textbf{for} $t = T, T-1, \dots, 1$: 
\State \quad $x \gets x - \text{Sharpened\_Denoise}(x, t, \text{denoise\_fn}, \delta, \alpha, t_{\text{threshold}})$
\end{algorithmic}
\end{algorithm}

\subsection{Estimating Laplacian in Higher Dimensions}

In a 1D distribution, the second derivative of the score can be computed by applying small perturbations in the forward and backward directions. In higher dimensions, this generalizes to the Laplacian, which represents the trace of the Hessian of the score. For instance, in 2D, the Laplacian can be approximated by applying perturbations along both dimensions, requiring four forward calls.
\[
\text{Laplacian}_{2D}(f) \approx 
\frac{
 f(x+\delta, y) + f(x-\delta, y) + f(x, y+\delta) + f(x, y-\delta) - 4 f(x, y)
}{\delta^2}
\]
However, this direct extension becomes computationally expensive for high-dimensional inputs such as images. To overcome this, we employ a finite-difference variant of the Hutchinson trace estimator, enabling efficient Laplacian estimation for each pixel. Crucially, the score function \( s(\mathbf{x}) \in \mathbb{R}^d \) is a vector unlike in the 1D setting; its output consists of \( d \) scalar components \( s_k(\mathbf{x}) \), each representing the score for a single pixel. The Hessian of the entire score function is therefore a third-order tensor. However, for our sharpening method, we require only the Laplacian for each output component \( s_k \)—that is, the trace of the Hessian \( \mathbf{H}_k \) of each scalar function \( s_k(\mathbf{x}) \). The Hutchinson identity states that for any such Hessian matrix \( H_{k} \in \mathbb{R}^{d \times d} \)

\begin{equation}
\mathrm{tr}(H_k) = \mathbb{E}_v \big[ v^\top H_k v \big],
\label{eq:E_k}
\end{equation}

where $v$ is a random Rademacher vector (i.e., each entry independently takes values $\pm 1$ with equal probability). $v^\top H_k v$ in Eq. \ref{eq:E_k} denotes the second directional derivative of \( s_k \) along the direction $v$. Using a central finite-difference approximation, it can be estimated as

\begin{equation}
v^\top H_k v \approx \frac{s_k(x + \delta v) - 2 s_k(x) + s_k(x - \delta v)}{\delta^2},
\label{eq:h_k}
\end{equation}
where $\delta$ is a small perturbation. Taking the expectation of Eq. \ref{eq:h_k} over Rademacher vectors $v$, the Laplacian of $s_k$ can be given as
\begin{equation}
Laplacian_k = \mathrm{tr}(H_k) \approx \mathbb{E}_v \Bigg[ \frac{s_k(x + \delta v) - 2 s_k(x) + s_k(x - \delta v)}{\delta^2} \Bigg].
\label{eq:L_k}
\end{equation}

This computation is performed for all pixels simultaneously by evaluating the score \(s(x\pm\delta v)\). This yields a vector \( L(\mathbf{x}) \in \mathbb{R}^d \) where the \( k \)-th entry is the Laplacian of \( s_k \). This vector \( L(\mathbf{x})\) will serve the function of  \( Y_j'' \) in Eq. \ref{eq:r_j}. We replace the second derivative in Algorithm \ref{alg:score_sharpening_1d} with the  \( L(\mathbf{x})\) for images.  This method allows us to efficiently estimate the Laplacian with significantly fewer function evaluations compared to computing second derivatives along each input dimension explicitly. The algorithm for sampling in images, including hyperparameter choices, is detailed in ~\ref{app:alg2}

\section{Experiments}
\label{headings}

\subsection{1D and 2D Synthetic Data}

We quantitatively evaluate our sharpening method on synthetic 1D and 2D distributions to measure its ability to reduce hallucinations while preserving distributional fidelity. We train a DDPM using a linear noise schedule over 1000 timesteps for 1000 epochs on the denoising objective. Hallucinated samples are defined as those lying beyond a predefined distance threshold from the modes of the training distribution. We report the Intermodulation Count (IM Count), which counts such inter-mode samples, and the L1 norm between the generated and target distributions at the end of 1000 epochs to assess overall fidelity.

For our sharpening method, we use a timestep threshold of \( t_{\text{threshold}} = 50 \), a perturbation scale of \( \delta = 0.1 \) (1D) and \( \delta = 0.05 \) (2D), and a sharpening strength of \( \alpha = 0.01 \) (1D) and \( \alpha = 0.0025 \) (2D). The values for \( \alpha \) were chosen based on the empirical ratio between the score and its second derivative.
\begin{figure}[H]
    \centering
    \includegraphics[width=1\linewidth]{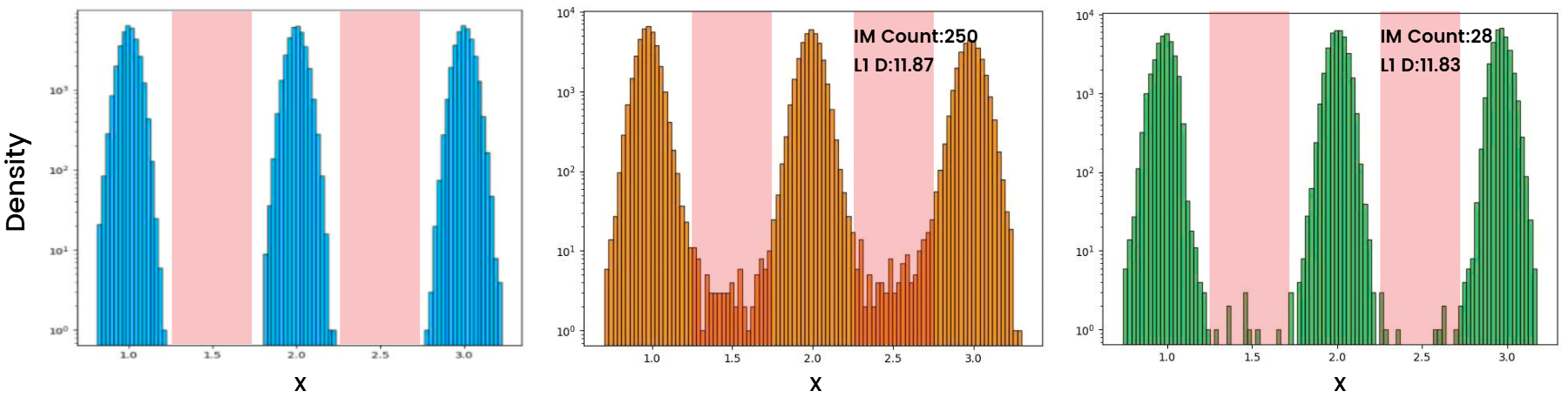}
    \caption{Histograms of the true 1D distribution (blue) and generated distributions using the vanilla (orange) and sharpened (green) scores for 100,000 samples. Sharpening the score reduces the number of samples generated in inter-mode regions.}
    \label{fig:1d_distribution}
\end{figure}

\begin{figure}[t]
    \centering
    \begin{minipage}{0.54\textwidth}
        \centering
        \includegraphics[width=1\linewidth]{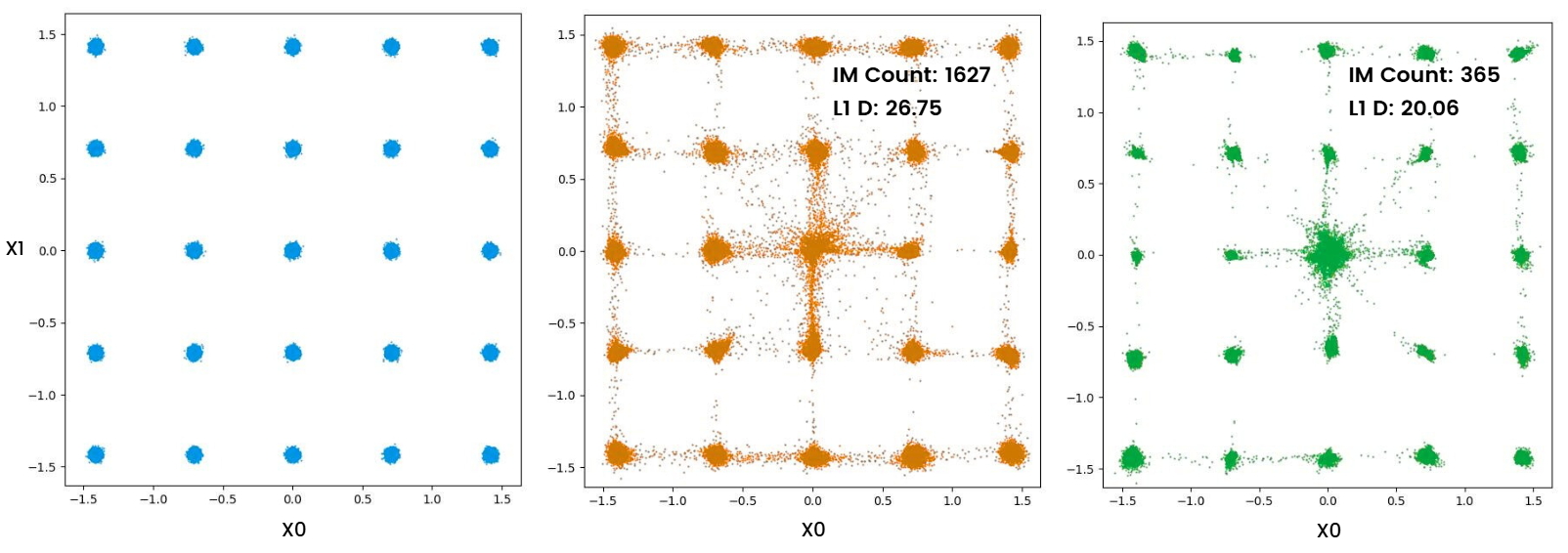}
        \caption{A Scatter Plot of the true 2D data distribution for 100,000 samples is shown in blue. Samples generated by the vanilla model (orange) exhibit grid-like interpolation artifacts between modes, which are reduced in samples generated using our sharpened score (red).}
        \label{fig:2d-distributions}
    \end{minipage}%
    \hfill
    \begin{minipage}{0.44\textwidth}
        \centering
        \captionof{table}{Image Quality Distribution: Vanilla vs. Sharpened Score for 1000 images. We see hallucinated images reduce by 3.9 percentage points and blank images increase by 4 percentage points.}
        \label{tab:image-quality-table}
        \resizebox{1\linewidth}{!}{
        \begin{tabular}{lcc}
            \toprule
            Image Type & Vanilla (\%) & Sharpened (\%) \\
            \midrule
            Hallucinated   & 6.00   & 2.10  \\
            Unknown Shapes & 0.60   & 0.60  \\
            Blank Images   & 1.30   & 5.30  \\
            Good Images    & 92.10  & 92.00 \\
            \bottomrule
        \end{tabular} }
    \end{minipage}
\end{figure}

After applying the sharpened score during inference, we observe a significant decrease in the IM Count, indicating a reduction in hallucinated samples. For instance, in the 1D case (Figure \ref{fig:1d_distribution}), the IM Count decreased from 250 to 28, while in 2D (Figure \ref{fig:2d-distributions}), it reduced from 1627 to 365. Crucially, the L1 norm either remains unchanged or shows a slight reduction, demonstrating that the core structure of the distribution is preserved while artifacts are mitigated.

\subsection{Rule-Based Toy Dataset: Shapes}

For image-based evaluation, we use the rule-based Shapes dataset introduced by \cite{aithal2024understandinghallucinationsdiffusionmodels} to quantitatively measure hallucinations. The training set always contains a single instance of each type of polygon—triangle, square, and pentagon—without repetition of the same shape. An image is classified as: 
\textbf{Good Image:} Contains one or more polygons with no repetition among the three valid shapes.
\textbf{Hallucinated Image:} Contains more than one polygon of the same shape (repetition).
\textbf{Unknown Shape:} Contains a polygon that is not a triangle, square, or pentagon.
\textbf{Blank Image:} Contains no shape and is predominantly noisy.

\begin{figure}[h!]
    \centering
    \begin{subfigure}[b]{0.33\linewidth}
        \centering
        \includegraphics[width=\linewidth]{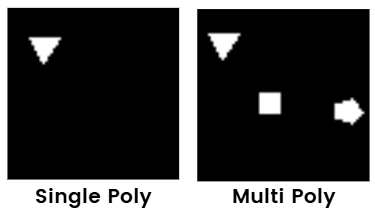}
        \caption{}
        \label{fig:good_samples}
    \end{subfigure}
    \begin{subfigure}[b]{0.5\linewidth}
        \centering
        \includegraphics[width=\linewidth]{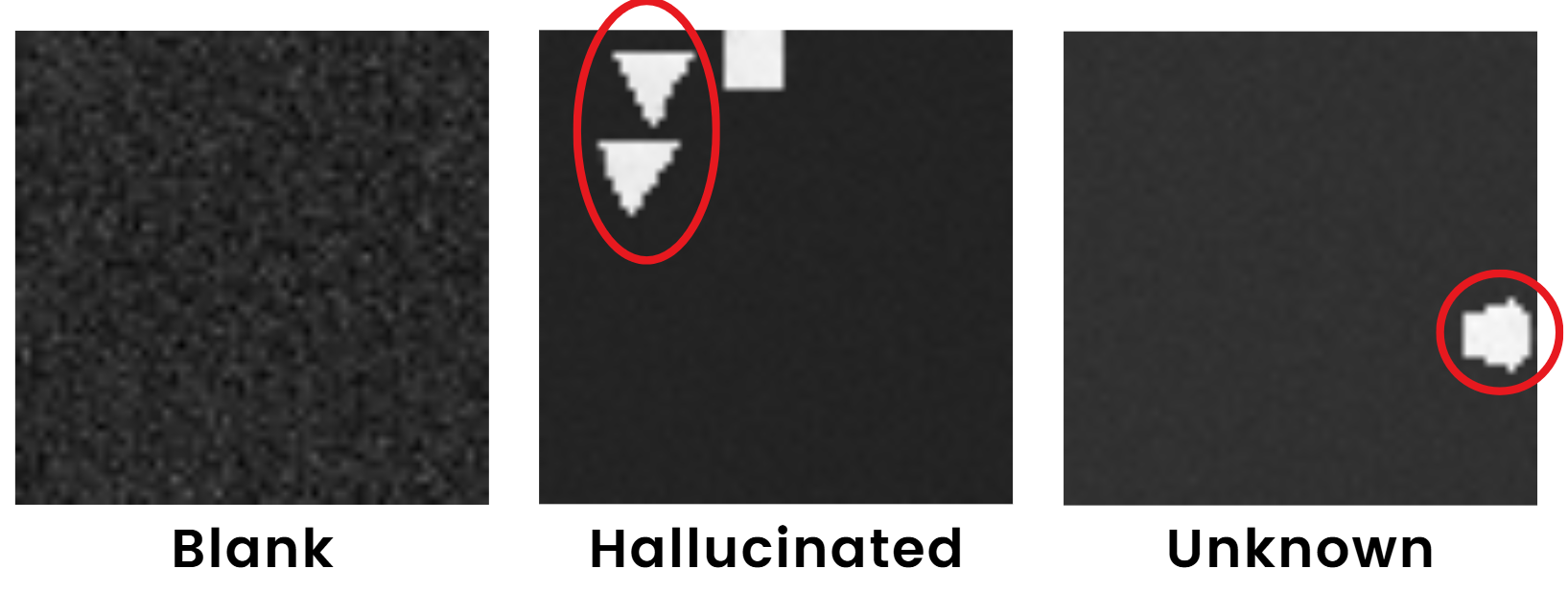}
        \caption{}
        \label{fig:bad_samples}
    \end{subfigure}
    \caption{(a) Samples on the true data manifold of Shapes Dataset, (b) Samples generated by the diffusion model, which are not part of the true data manifold.}
    \label{fig:placeholder}
\end{figure}

We train a unconditional DDPM with UNET architecture using a cosine noise scheduler similar to ADM \cite{dhariwal2021diffusionmodelsbeatgans} over 1000 timesteps for 50 epochs.
After applying our sharpening method during inference, we observe a strong reduction in mode interpolation hallucinations as shown in Table \ref{tab:image-quality-table}. The proportion of hallucinated images dropped from 6.0\% to 2.1\%, affirming that sharpening effectively guides samples away from the inter-mode regions. However, this intervention also resulted in a corresponding increase in blank images from 1.3\% to 5.3\%, suggesting that while sharpening pushes samples away from inter-mode regions, it does not subsequently guide them toward the true data manifold as effectively as it does in 1D or 2D settings. The proportions of good and unknown-shape images remained stable, indicating that the core generative capability is preserved.

\begin{figure}[h!]
    \centering
    \includegraphics[valign=m,width=0.11\linewidth]{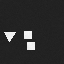}
    \includegraphics[valign=m,width=0.15\linewidth]{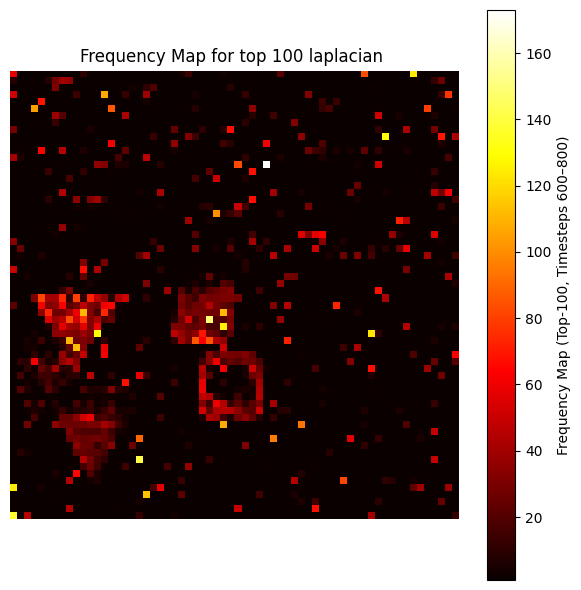}
    \includegraphics[valign=m,width=0.41\linewidth]{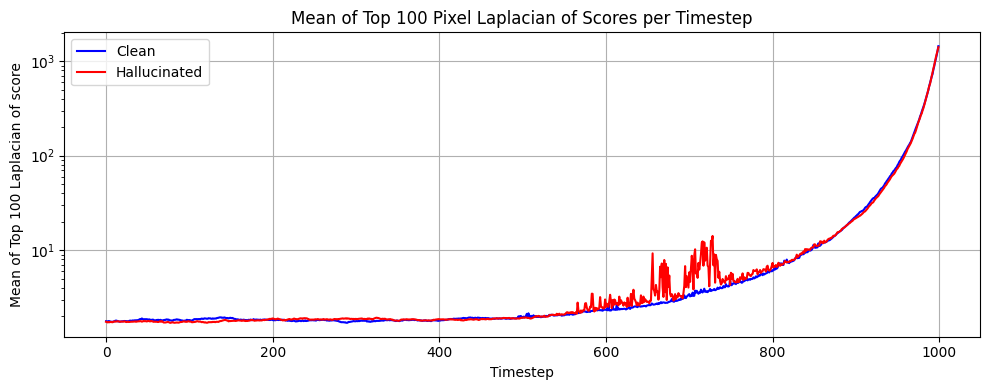}
    \includegraphics[valign=m,width=0.11\linewidth]{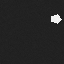}
    \includegraphics[valign=m,width=0.15\linewidth]{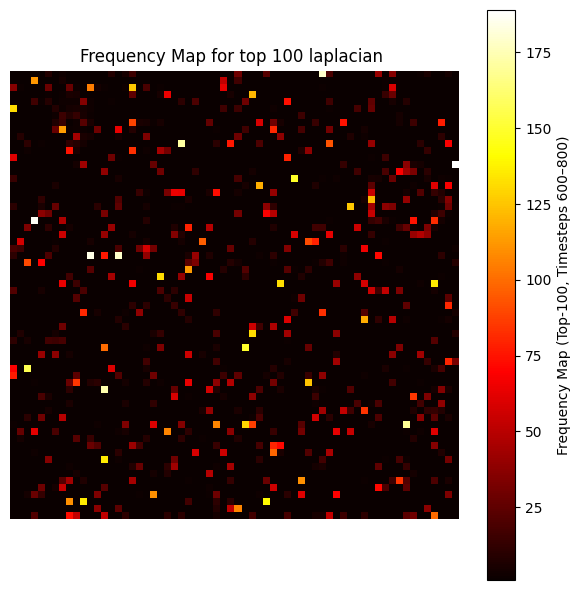}
    \caption{\textbf{Left}: Frequency map of top 100 absolute Laplacian magnitudes (timesteps 600–800) for a hallucinated image (red curve, middle). \textbf{Middle}: Mean of the top 100 absolute Laplacian magnitudes over time. \textbf{Right}: Frequency map for a clean image (blue curve, middle).}
    \label{fig:Laplacian}
\end{figure}

On plotting the mean of the top 100 absolute Laplacian magnitudes in \( |L(\mathbf{x})| \) corresponding to each pixel (Figure \ref{fig:Laplacian}, Middle), we observed a clear peak in the mean between 600 and 800 timesteps (in ascending order of sampling) for the majority of hallucinated samples. Interestingly, this peak was not exclusive to images with final hallucinations; it also appeared in a subset of ultimately clean samples that exhibited transient artifacts— hallucinated structures that emerged mid-sampling but were later corrected. Furthermore, when we mapped the pixels corresponding to these top 100 values on a frequency map(Figure \ref{fig:Laplacian}, Left \& Right), we found that these high values were associated with hallucinated shapes—either those appearing in the final image or those that emerged during the image trajectory but were absent in the final image.

\section{Conclusion and Future Work}

In this work, we introduced a novel, post-hoc sharpening technique for the score function using its Laplacian to reduce hallucinations. We derived an efficient method to approximate the high-dimensional Laplacian using a finite-difference variant of the Hutchinson estimator, making our method scalable to images. We provided an analysis linking the Laplacian of the score to uncertainty and hallucination during the sampling process, validated for 1D setting and the image dataset-Shapes.

Though the method is able to reduce hallucinated samples in the high dimensional Shapes dataset, the core limitation of this corrective approach is its destructive nature. Consequently, it is ideal for pruning errors but less suitable for open-ended discovery tasks like de novo protein generation, where the goal is to explore and refine uncertain regions into viable solutions. An additional limitation is its sensitivity to hyperparameters, such as the sharpening strength $\alpha$ and the perturbation scale $\delta$, which can be challenging to tune optimally. Future work will focus on integrating the Laplacian signal more adaptively into the sampling process to not only avoid artifacts but also proactively guide samples toward the data manifold. We will also explore methods to automate the selection of $\alpha$ and $\delta$ to improve the algorithm's practicality.


{
\small
\bibliographystyle{plain}
\bibliography{main}
}

\appendix

\section{Technical Appendices and Supplementary Material}

\subsection{Evolution of Score over the Timesteps}
\label{app:time}
\begin{figure}[H]
    \centering
    
    \begin{subfigure}[b]{0.24\textwidth}
        \centering
        \includegraphics[width=\linewidth]{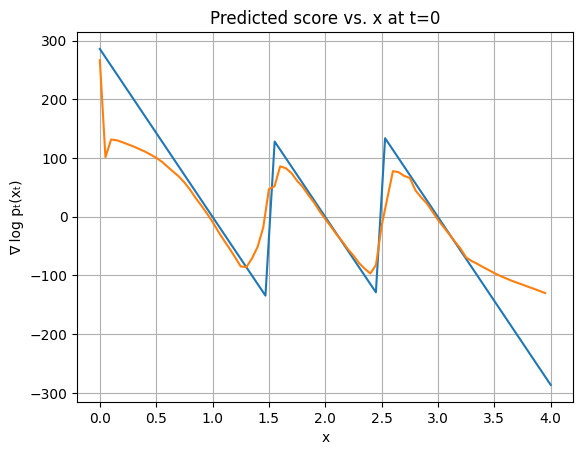}
        \caption{Sharpened t=0}
        \label{fig:sharp_t0}
    \end{subfigure}
    \begin{subfigure}[b]{0.24\textwidth}
        \centering
        \includegraphics[width=\linewidth]{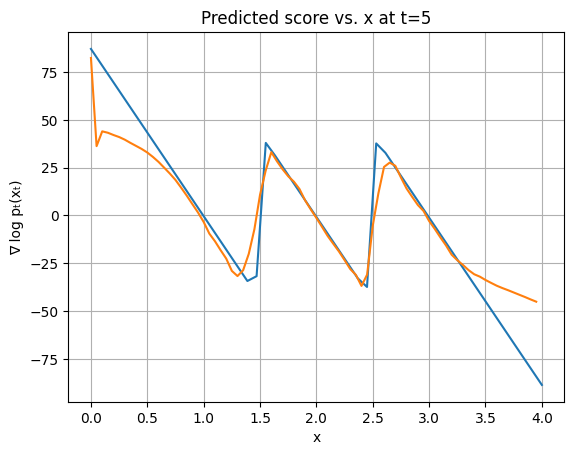}
        \caption{Sharpened t=5}
        \label{fig:sharp_t5}
    \end{subfigure}
    \begin{subfigure}[b]{0.24\textwidth}
        \centering
        \includegraphics[width=\linewidth]
        {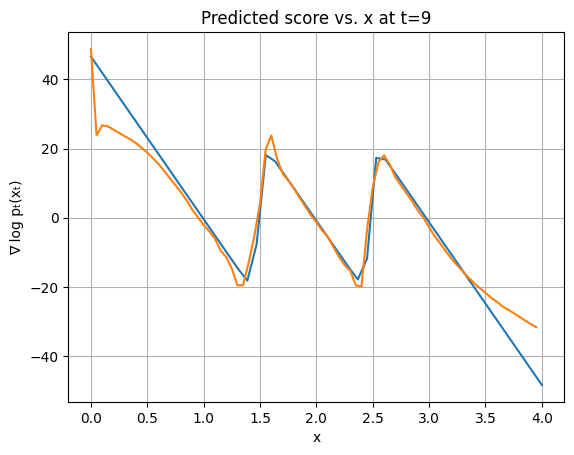}
        \caption{Sharpened t=10}
        \label{fig:sharp_t10}
    \end{subfigure}
    \begin{subfigure}[b]{0.24\textwidth}
        \centering
        \includegraphics[width=\linewidth]{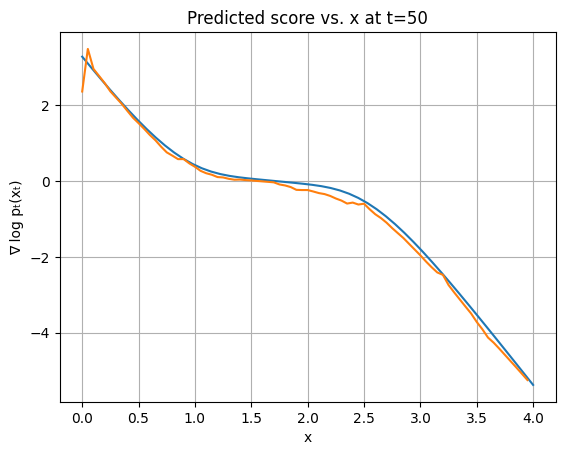}
        \caption{Sharpened t=50}
        \label{fig:sharp_t50}
    \end{subfigure}
    
    \begin{subfigure}[b]{0.24\textwidth}
        \centering
        \includegraphics[width=\linewidth]{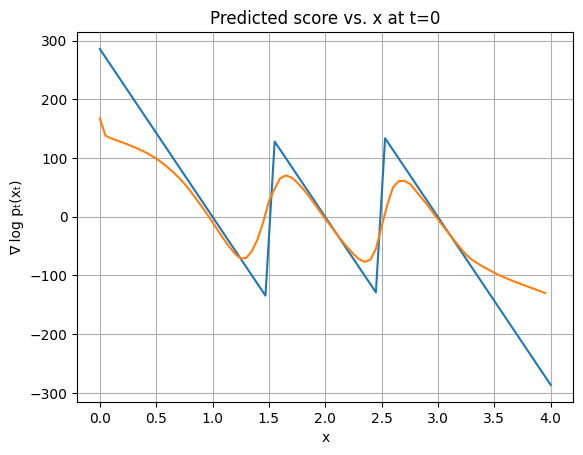}
        \caption{Vanilla t=0}
        \label{fig:vanilla_t0}
    \end{subfigure}
    \begin{subfigure}[b]{0.24\textwidth}
        \centering
        \includegraphics[width=\linewidth]{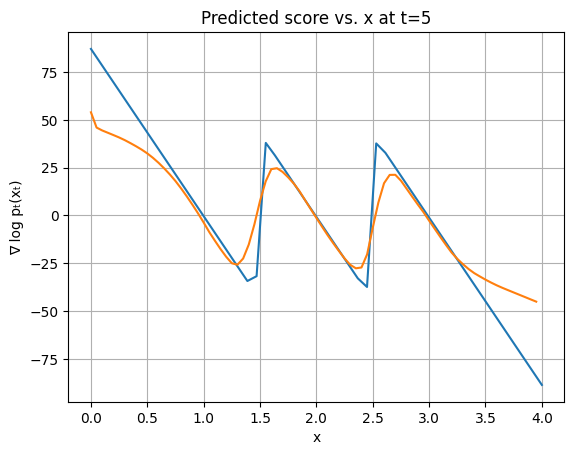}
        \caption{Vanilla t=5}
        \label{fig:vanilla_t5}
    \end{subfigure}
    \begin{subfigure}[b]{0.24\textwidth}
        \centering
        \includegraphics[width=\linewidth]{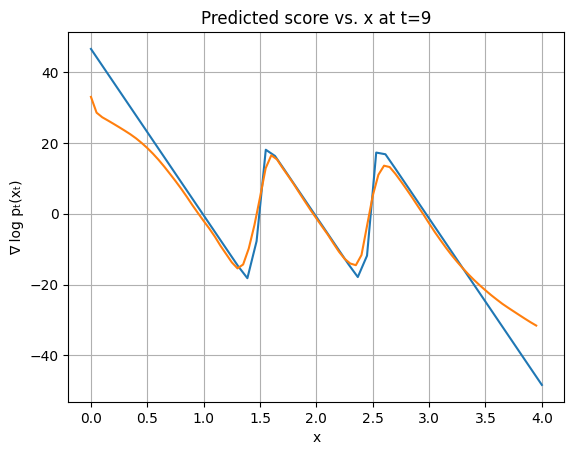}
        \caption{Vanilla t=10}
        \label{fig:vanilla_t10}
    \end{subfigure}
    \begin{subfigure}[b]{0.24\textwidth}
        \centering
        \includegraphics[width=\linewidth]{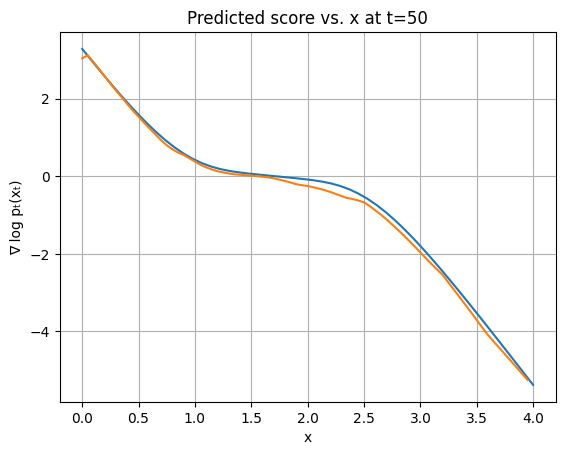}
        \caption{Vanilla t=50}
        \label{fig:vanilla_t50}
    \end{subfigure}
    
    \caption{Comparison of sharp vs vanilla scores across different timesteps (forward).}
    \label{fig:score_comparison}
\end{figure}

Figure~\ref{fig:score_comparison} illustrates the evolution of the score function in the 1D setting, comparing the learned score with and without sharpening. We observe that the sharpened score aligns more closely with the true score than the learned (vanilla) score up to approximately timestep 50 (corresponding to a late stage in the reverse diffusion where noise is low), effectively recovering sharper mode boundaries. Beyond timestep 50, however, the true score no longer exhibits pronounced peaks, and applying sharpening in this regime may introduce unnecessary artifacts rather than improving alignment.

\subsection{Laplacian sharpening Algorithm for Images}
\label{app:alg2}
For the Hutchinson estimation, $n_{samples}$ specifies the number of Rademacher vectors sampled to approximate the second derivative. In our experiments, we set $n_{samples}$= 3. Unlike in 1D or 2D, we found that applying score sharpening during early timesteps is ineffective. This is because the model primarily learns finer details in the early timesteps, whereas hallucinations tend to appear at later timesteps. To address this, we introduce a timestep range for sharpening, with a lower bound to avoid early timesteps. Specifically, we used the range (200, 400) in the forward diffusion convention, or (600, 800) in the sampling convention, as inferred from Figure~\ref{fig:Laplacian}, Middle. We use $\delta = 0.05$ and $\alpha = 0.05$ with $\alpha$ being inferred based on the observed ratio between the score and its second derivative between 600 and 800 timesteps.

\begin{algorithm}[H]
\caption{Score Sharpening with Rademacher Perturbations}
\begin{algorithmic}[1]
\Function{Sharpened\_Denoise}{$x, t, t_{\text{low}}, t_{\text{high}}, \text{denoise\_fn}, \delta, \alpha, n_{\text{samples}}$}
    \State $f_x \gets \text{denoise\_fn}(x, t)$ \Comment{estimates noise vector $\epsilon$ at timestep $t$}
    \If{$t_{\text{low}} < t < t_{\text{high}}$} 
        \Comment{$(t_{\text{low}}, t_{\text{high}})$: timestep range}
        \State $\text{Laplacian\_total} \gets 0$ 
        \For{$i = 1$ \textbf{to} $n_{\text{samples}}$}
            \State $r \gets \text{Rademacher}(x)$ \Comment{Random ±1 perturbation per element}
            \State $h_i \gets \delta \cdot r$ \Comment{$\delta$: perturbation size}
            \State $f_x^{+} \gets \text{denoise\_fn}(x + h_i, t)$
            \State $f_x^{-} \gets \text{denoise\_fn}(x - h_i, t)$
            \State $\text{Laplacian\_total} \gets \text{Laplacian\_total} + \frac{f_x^{+} - 2 f_x + f_x^{-}}{h_i^2}$
        \EndFor
        \State $\text{Laplacian} \gets \text{Laplacian\_total} / n_{\text{samples}}$
        \State $f_x \gets f_x - \alpha \cdot \text{Laplacian}$ \Comment{$\alpha$: sharpening strength}
    \EndIf
    \State \Return $f_x$
\EndFunction
\end{algorithmic}
\end{algorithm}


\end{document}